\documentclass[twoside,11pt]{article}

\usepackage[specialissue, accepted]{melba}

\usepackage{amsmath,amsfonts}
\melbaid{2024:008}  % This is provided upon by the publishing editor
\doi{https://doi.org/10.59275/j.melba.2024-8g8b}
\melbaauthors{Fischer, Kiechle, Lang, Peeken and Schnabel}  % Note: this one is also used to set the pdf 'authors' metadata
\volume{2}
\firstpageno{798}  % Communicated by the publishing editor
\melbayear{2024}  % The publication year
\datesubmitted{01/2024}  % Date submitted to MELBA: mm/yyyy
\datepublished{06/2024}  % Today's date: mm/yyyy

\melbaspecialissue{MICCAI 2023 Lymph Node Quantification Challenge}
\melbaspecialissueeditors{Steve Pieper, Erik Ziegler, Tawa Idris, Bhanusupriya Somarouthu, Reuben \\
Dorent, Gordon Harris, Ron Kikinis}

\ShortHeadings{Mask the Unknown}{Fischer, Kiechle, Lang, Peeken and Schnabel}

\title{Mask the Unknown: Assessing Different Strategies to Handle Weak Annotations in the MICCAI2023 Mediastinal Lymph Node Quantification Challenge}

\author{\firstname Stefan M. \surname Fischer$^{1, 2, 3, 4}$ \email stefan.mi.fischer@tum.de \\  % start right after \author{, or there will be an extra space
	\AND
	\name Johannes Kiechle$^{1, 2, 3, 4}$ \email johannes.kiechle@tum.de \\
 	\AND
	\name Daniel M. Lang$^{1, 3}$ \email daniel.lang@tum.de \\
 	\AND
	\name Jan C. Peeken$^{2}$ \email jan.peeken@tum.de \\
  	\AND
	\name Julia A. Schnabel$^{1, 3, 4, 5}$ \email julia.schnabel@tum.de \\
    \AND 
	\addr 1: School of Computation, Information and Technology, Technical University Munich, Germany \\
 	\addr 2: Department of RadioOncology, Klinikum rechts der Isar, Technical University Munich, Germany \\
  	\addr 3: Institute of Machine Learning in Biomedical Imaging, Helmholtz Munich, Germany \\
    \addr 4: Munich Center of Machine Learning (MCML), Germany \\
    \addr 5: School of Biomedical Engineering and Imaging Sciences, King's College London, UK \\
}

\begin{document}

% top matter
\maketitle

% abstract
\begin{abstract}%   <- trailing '%' for backward compatibility of .sty file
    Pathological lymph node delineation is crucial in cancer diagnosis, progression assessment, and treatment planning. The MICCAI 2023 Lymph Node Quantification Challenge published the first public dataset for pathological lymph node segmentation in the mediastinum. As lymph node annotations are expensive, the challenge was formed as a weakly supervised learning task, where only a subset of all lymph nodes in the training set have been annotated. For the challenge submission, multiple methods for training on these weakly supervised data were explored, including noisy label training, loss masking of unlabeled data, and an approach that integrated the TotalSegmentator toolbox as a form of pseudo labeling in order to reduce the number of unknown voxels. Furthermore, multiple public TCIA datasets were incorporated into the training to improve the performance of the deep learning model. Our submitted model achieved a Dice score of 0.628 and an average symmetric surface distance of 5.8~mm on the challenge test set. With our submitted model, we accomplished third rank in the MICCAI2023 LNQ challenge. A finding of our analysis was that the integration of all visible, including non-pathological, lymph nodes improved the overall segmentation performance on pathological lymph nodes of the test set. Furthermore, segmentation models trained only on clinically enlarged lymph nodes, as given in the challenge scenario, could not generalize to smaller pathological lymph nodes. The code and model for the challenge submission are available at \url{https://gitlab.lrz.de/compai/MediastinalLymphNodeSegmentation}.

\end{abstract}

% keywords

% Introduction (or first section)
\section{Introduction}

In the following section, we introduce the "Mediastinal Lymph Node Quantification: Segmentation of Heterogeneous CT Data" (LNQ2023) challenge, and discuss related work.
\subsection{Motivation}

Lymph nodes (LNs) are small anatomical structures scattered throughout the body. Based on their location, they are grouped into various LN stations according to established definitions, such as those from the International Association for the Study of Lung Cancer (IASLC) \citep{rusch2009iaslc}. During cancer progression, the tumor grows, and cancer cells spread into nearby anatomical structures, developing into metastasis and radically increasing the severity of the disease. Infiltration of cancerous tissue into LNs may lead to enlargement of those LNs. Thus, assessing metastatic LNs is a critical factor for initial diagnosis, tumor staging, and treatment planning. The conventional criteria for quantifying lymph node size is based on Response Evaluation Criteria In Solid Tumours (RECIST) guidelines \citep{eisenhauer2009new}. Enlarged LNs are defined as those whose shortest diameter exceeds 10~mm on an axial CT slice. Medical professionals solely rely on unidirectional or bidirectional measurements on a single axial slice of just one or a few LNs, introducing limitations in capturing the full extent of abnormalities \citep{guo2022thoracic}.

However, studies indicate that solely relying on the feature of shortest LN diameter for malignancy assessment yields recall rates of only 60\%-80\% in lung cancer patients \citep{yan2023anatomy}. The facts that accessing the status of LNs is a complex, time-consuming task and that the shortest diameter is a limiting metric highlights the necessity for accurate segmentation in three dimensions to comprehensively evaluate lymph node disease \citep{guo2022thoracic}. Furthermore, precise delineation of all tumorous regions is particularly crucial in radiation therapy, where irradiating metastatic areas such as LNs impacts patient outcomes significantly \citep{chapet2005ct}. Thus, automated segmentation not only holds promise for reducing the inter and intra-observer variability, but also reduces the task-related working time.

%%%%%%%%%%%%%%%%%%%%%%% Introduce LNQ Challenge %%%%%%%%%%%%%%%%%%%%%%%%%%%%%%
The objectives of the LNQ2023 challenge are twofold, each aiming to address critical aspects in the field of lymph node identification and segmentation \citep{roya_khajavibajestani_2023_7844666}. The primary goal of the challenge was to establish a benchmark for the detection and segmentation of mediastinal lymph nodes. The mediastinum, located between the lung lobes, poses a particular challenge due to the presence of ten or more lymph nodes, often with three or more enlarged nodes exceeding 10~mm in diameter. The secondary goal focuses on the exploration and application of weakly supervised learning techniques in the scenario of LN segmentation. Given the time-consuming nature of manual annotation and the possible presence of several pathological LN instances, there is a lack of pre-existing fully annotated pathological LN datasets. This setting aligns with the current growing interest in the medical imaging community in harnessing weak annotations \citep{kemnitz2018combining, petit2018handling, zhou2019prior, shi2021marginal, dong2022towards, ulrich2023multitalent}.

%%%%%%%%%%%%%%%%%%%%%%%%%%%%%%%%%%%%%%%%%%%%%%%%%%%%%%%%%%%%%%%%%%%%%%%%%%%
% Related works
%%%%%%%%%%%%%%%%%%%%%%%%%%%%%%%%%%%%%%%%%%%%%%%%%%%%%%%%%%%%%%%%%%%%%%%%%%%
% Make sure to put your work into context and include apporpriate citations.
% We do not have limits on citation counts.

\subsection{Mediastinal Lymph Node Segmentation}

The first public dataset for mediastinal LN segmentation was introduced by \cite{roth2014new, roth2015tcia}, offering segmentation annotations of enlarged lymph nodes. This task presents unique challenges due to reduced contrast compared to axillary and pelvic nodes \citep{nogues2016automatic}. Early methods often relied on machine learning approaches that incorporated handcrafted features and manual region-of-interests (ROI) \citep{aerts2014decoding, liu2014mediastinal, liu2016mediastinal, oda2017hessian, oda2017automated, roth2014new}. Notably, \cite{roth2014new} introduced the first neural network (NN) based method, specifically for LN detection after an ROI proposal step, reducing false positives and marking a shift in methodology.

Building upon this, subsequent works explored various NN-based methods operating directly on full CT volumes \citep{bouget2019semantic, bouget2021mediastinal, guo2022thoracic, yan2023anatomy, iuga2021automated}.

To enhance segmentation accuracy and robustness, the inclusion of anatomical key regions to guide the segmentation has been explored \citep{bouget2021mediastinal, oda2018dense, bouget2019semantic}. The integration of lymph node station information has been a focal point, initially introduced by \cite{liu2014mediastinal}. The idea of mapping lymph node stations, according to IASLC guidelines, has been extended by various methods, including NN-based approaches. \cite{guo2022thoracic} proposed dedicated encoders for different LN stations, and \cite{yan2023anatomy} introduced a station-stratified LN detector, emphasizing the importance of incorporating station information during training \citep{liu2014mediastinal, guo2022thoracic, yan2023anatomy}.

%%%%%%%%%%%%%%%%%%%%%%%%%%%%%%%%%%%%% Weakly supervised learning %%%%%%%%%%%%%%%%%%%%%

\subsection{Weakly Supervised Learning}

Weakly supervised learning has recently become a popular topic in medical imaging research, as most datasets only provide annotations for one single or few structures of interest. Combining multiple datasets results in a partially supervised learning setting, as datasets come with full supervision of a few classes. It forms a special case of weak supervision. Several research groups have concentrated on such partially supervised learning approaches, aiming to enhance model performance by aggregating information from multiple partially labeled datasets during training \citep{kemnitz2018combining,petit2018handling,zhou2019prior,shi2021marginal,dong2022towards,ulrich2023multitalent}. Some works focus on masking the loss of missing classes in the current training label \citep{ulrich2023multitalent,dong2022towards}. Furthermore, different approaches to constrain the loss functions were motivated by mutual exclusion of classes or implemented by merging classes to superclasses \citep{kemnitz2018combining,shi2021marginal,ulrich2023multitalent,petit2018handling}.

However, there is only limited work outside the partially supervised setting. In the computer vision domain, missing labels are often treated as background, a method viewed as a simplistic form of dealing with noisy labels. This approach is effective when the pixels of missing classes constitute a significantly smaller portion of images compared to background pixels \citep{dong2022towards}. 

A general strategy for training on weakly labeled datasets is masking the loss of voxels lacking annotations \citep{kemnitz2018combining}. Additionally, incorporating semi-supervised learning (SSL) has been explored to gain additional training feedback by mining unlabeled voxels \citep{zhou2019prior, petit2018handling}. 

%%%%%%%%%%%%%%%%%%%%%%%%%%%%%% missing voxel annotations %%%%%%%%%%%%%%%%%%%%%%%

Most works in SSL focus on the classification or segmentation of completely unlabeled images. In contrast, the LNQ2023 challenge is given as an incomplete pixel-level label scenario in which one or multiple foreground instances are annotated. \cite{nguyen2020learning} have explored this particular scenario. They focused on two distinct scenarios: speech balloon segmentation in comics and cell segmentation in medical imaging. In both cases, the incomplete annotation is generated using an automatic extraction method. The learning of background information is facilitated by a small set of background voxels chosen as direct neighbors of known foreground instances. By applying an SSL technique, they were able to learn the representation of foreground and background.

\subsection{Problem Setting and Contributions}

The LNQ2023 challenge goal was to segment all pathological mediastinal LNs from thorax CT volumes, while the given training labels only covered some instances of the foreground. Those instances, furthermore, were only covering enlarged pathological LN components. The problem setting was a weakly or semi-supervised learning task with incomplete pixel-level annotations. As the challenge was an open challenge, it was allowed to use the public CT Lymph Nodes dataset from TCIA \citep{clark2013cancer, roth2015tcia}, providing a set of CT volumes in which enlarged LNs were fully annotated. Our challenge strategy was to develop a supervised training strategy, handling the incomplete pixel-level labels of the LNQ2023 challenge data. Furthermore, we integrated additional public data and the TotalSegmentator \citep{wasserthal2023TotalSegmentator} to gain performance improvements.

Our main contributions are summarized as follows:
\begin{enumerate}
    \item Starting from a fully annotated dataset of image volumes, namely the public TCIA Lymph Nodes dataset, we implemented different strategies to integrate the additional incomplete pixel-level labeled data, as given in the LNQ2023 challenge, into the training process. Those strategies were noisy label training, loss masking, and wrapping each foreground instance with a background shell.
    \item We applied the public toolbox TotalSegmentator to identify anatomical structures and, by exclusion, set those to background class voxels. We refer to this as TotalSegmentator Pseudo Labeling. 
    \item We explored the effect of integrating different public datasets, namely TCIA CT Lymph Nodes, an annotation-refined version of CT Lymph Nodes, TCIA NSCLC-Radiomics, TCIA NSCLC-Radiogenomics, and TCIA NSCLC-Radiomics-Interobserver on the downstream performance.
    \item Furthermore, we performed experiments on the impact of adding all visible, potentially non-pathological lymph nodes to the model training on the overall segmentation performance and the performance regarding the lymph node's shortest diameter.  
\end{enumerate}

% A methodological, model, or similar section often comes here.
\section{Integration of Weakly Annotated Data}
\label{methods}

\begin{figure}
            \centering
    		\includegraphics[width=\linewidth]{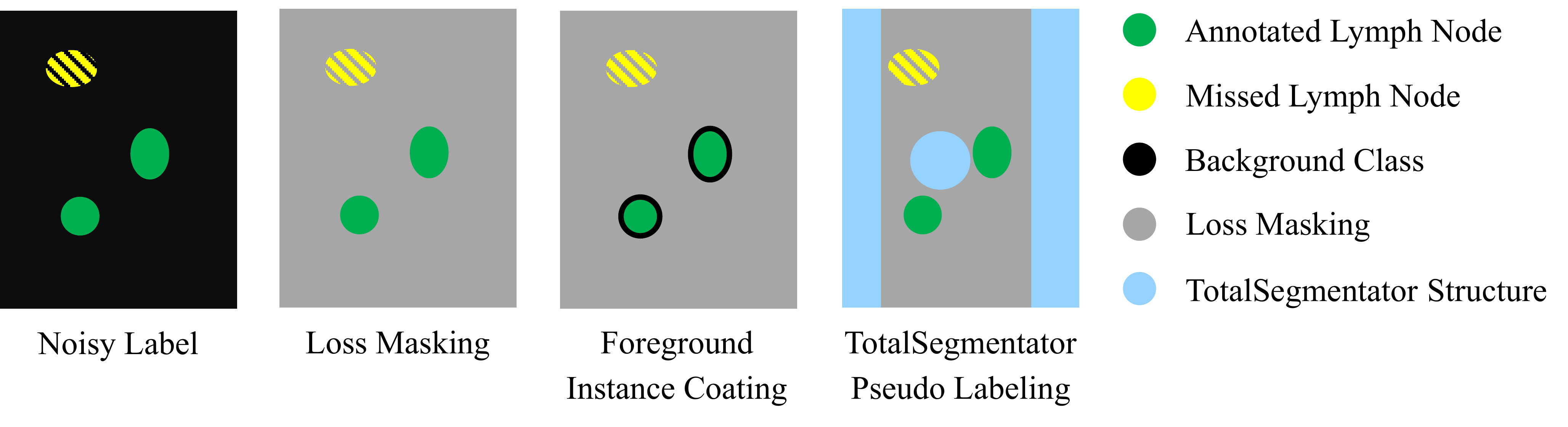}

            \caption{Sketch of different strategies to handle weakly annotated data in our analysis. The missed lymph node instance is incorrectly set to the background class in the noisy label training. For loss masking, foreground instance coating, and TotalSegmentator Pseudo Labeling, the missing instance is removed from the training process by loss masking.}
            \label{Fig:strategy_sketch}
    	\end{figure}

 In the following section, we present the different strategies to integrate the weakly annotated data into the training process, shown in Figure~\ref{Fig:strategy_sketch}. We cropped the input CT volumes with the help of the TotalSegmentator, a deep learning-based toolbox capable of segmenting 104 different anatomical structures on CT volumes \citep{wasserthal2023TotalSegmentator}. Using the toolbox, we created a bounding box of the lung lobes, to which the CT volume was cropped. The resulting image volume then contained the full mediastinum, leading to improved computational efficiency. As a segmentation network, we used the nnUNet, a fully self-configuring segmentation pipeline \citep{isensee2021nnu}. 
          
    \paragraph{Noisy Label}

        The unlabeled voxels of the LNQ2023 training set were set to the background class as a form of noisy labels. Consequently, unlabeled LN instances were naïvely set to background. The foreground only consisted of the expert-annotated foreground instances.

    \paragraph{Loss Masking}

        Another simple strategy to include the weakly labeled image volumes in the training procedure was to mask out regions without class annotation. For such voxels, the loss was set to zero so that only labeled voxels contributed to the learning process.

    \paragraph{Foreground Instance Coating}

        Given foreground LN instances in the weakly annotated LNQ2023 training data, we followed the approach of \cite{nguyen2020learning} and set voxels neighboring the foreground instances to the background class. We implemented that by running the morphological operator binary dilation. In this way, each foreground component was embedded in a hull of background voxels.

    \paragraph{TotalSegmentator Pseudo Labeling}

        All 104 output structures of the TotalSegmentator, covering organs, vessels, bones, and muscles, by definition, should not contain any LNs. We exploited this fact to constrain our problem setting. We utilized the output classes to annotate unlabeled voxels as background if classified as foreground by the TotalSegmentator. During that process, we skipped overwriting expert annotated LN instances in the LNQ2023 training set. This strategy is especially beneficial in the mediastinum, as the TotalSegmentator covers many known anatomical structures located within the mediastinum.

\section{Experiments}
\label{Experiments}

In this section, we describe the ablation study we performed to select the building blocks of the final model submission. For this purpose, we added various training components from the network training, such as extra training data or different training strategy changes, to analyze their contribution to the model's performance on the test set.

    \subsection{Data}

            \begin{table}
            \centering
            \begin{tabular}{l | r | r | r | r}
            Dataset & Volumes & Labeled LNs & Labeled Enlarged LNs & Fully Labeled \\
            \hline \hline
            LNQ2023 train set & 393 & 558 & 512 & No  \\
            LNQ2023 test set & 100 & 845 & 289 & Yes  \\
            CT Lymph Nodes & 90 & 294 & 244 & Yes \\
            Bouget Refinements & 90 & 1403 & 414 & Yes \\
            NSCLC datasets & 585 & 0 & 0  & No \\
           \end{tabular}
           \caption{Lymph node statistics per dataset. A lymph node component is considered enlarged if its shortest diameter is equal to or greater than 10~mm.}
           \label{tab:LNs_Stats_Per_Dataset}
        \end{table}
        
        \begin{figure}
            \centering
            \includegraphics[width=\textwidth]{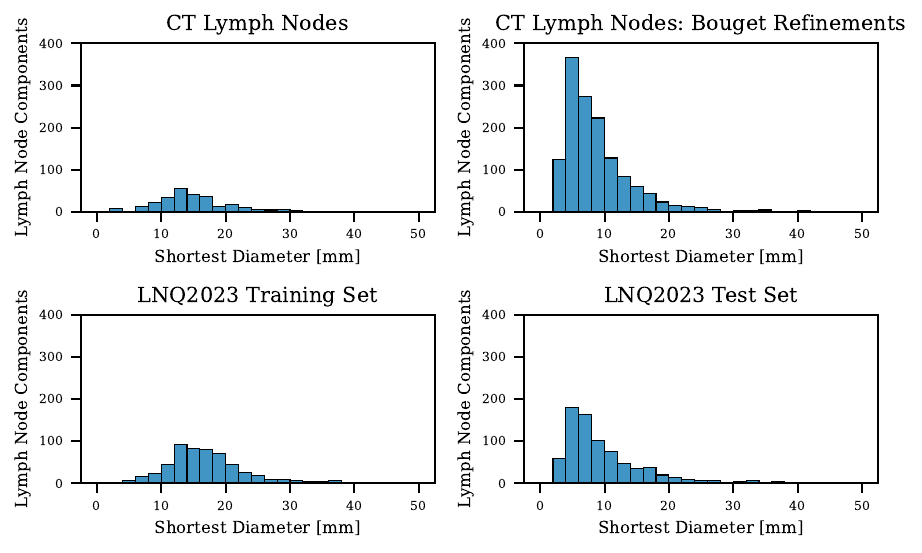}
            \caption{Histograms of shortest diameter of lymph node components in TCIA CT Lymph Nodes dataset, refined annotations by \cite{bouget2019semantic} and the LNQ2023 training and test set.}            
            \label{Figure:LNs_Stats_Per_Dataset}
        \end{figure}

        We included three different source datasets for training the segmentation model. The LNQ2023 training data consists of 393 weakly annotated thorax CT volumes of patients suffering from various cancer types, including breast cancer, leukemia, lung cancer, and others.

        Another dataset used is the TCIA CT Lymph Nodes dataset introduced by \cite{roth2015tcia}. It contains 90 thorax CT volumes with fully annotated mediastinal LNs. Here, only clinically enlarged LN instances are annotated. The TCIA CT Lymph Nodes dataset was refined in the work of \cite{bouget2019semantic}. They updated the existing annotations and integrated all visible LNs without separating them into pathological or healthy LNs. 

        Furthermore, to increase the amount of data, we added three different lung cancer datasets, adding up to 585 CT volumes, to the training data. The first is a subset of the NSCLC-Radiogenomics dataset, consisting of 143 thorax CT volumes with tumor delineation \citep{gevaert2012non, nsclc_radiogenomics, bakr2018radiogenomic, clark2013cancer}. The other lung cancer dataset is the NSCLC-Radiomics containing 422 thorax CT volumes with tumor delineation \citep{aerts2014decoding,nsclc_radiomics,clark2013cancer}. Additionally, we integrated the NSCLC-Radiomics-Interobserver1 dataset by adding the 20 thorax CT volumes and the delineation of the tumor by the medical expert 1 to the data pool \citep{nsclc_radiomics_interobserver, clark2013cancer, aerts2014decoding}. Multiple individuals of the NSCLC datasets developed metastatic LNs documented in the patient's N-staging.

        The occurring LN instances per dataset were generated by a connected component analysis, and statistics of the datasets are specified in Table~\ref{tab:LNs_Stats_Per_Dataset}. Not all LN components in the CT Lymph Nodes dataset were considered enlarged. The Bouget refinements, which are refined annotations of the standard CT Lymph Nodes dataset, contained more than four times as many annotated components. The refined version had almost twice as many enlarged LN instances. LNQ2023 training set is not fully annotated, so the number of LNs per CT volume should be similar to the LNQ2023 test set. For the NSLC datasets, roughly half of the patients suffer from pathological thorax LNs, while at least 171 patients had spread into the mediastinal LNs. The histograms of the shortest LN component diameter per labeled dataset are shown in Figure~\ref{Figure:LNs_Stats_Per_Dataset}.

    \subsection{Ablation Study for Integration Strategy of Weakly Annotated Data}
        \label{sec:ablation_studies_of_supervised_baseline}

        We performed an ablation study to evaluate the various training strategy building blocks. The LNQ2023 challenge test set, containing 100 fully annotated samples, was used for evaluation. The Dice score and the average symmetric surface distance (ASSD) were computed for performance assessment. Initially, during the challenge, we used 20 samples of the CT Lymph Nodes dataset for validation of the method development, which we, therefore, omitted in the trained models of the ablation study.

       We used default nnUNet planning and training for all models. All different training strategies introduced in Section~\ref{methods} and various combinations of data for training were evaluated on the LNQ2023 test set:

        \begin{itemize}

            \item \textbf{Model 1}: Baseline segmentation model trained on 70 fully annotated samples of the TCIA CT Lymph Nodes dataset.  Thus, training annotations only included enlarged LN components. 
            \item \textbf{Model 2}: Model trained on data combination of TCIA CT Lymph Node and LNQ2023 training set. Unlabeled voxels were set to background class, so the model was trained in a noisy label fashion.
            \item \textbf{Model 3}: Instead of noisy label training here, loss masking was applied. The loss of the unlabeled voxels was set to zero so that it did not affect the training process. 
            \item \textbf{Model 4}: The LNQ2023 training data was preprocessed with foreground instance coating with a background margin of one voxel for each hull.             
            \item \textbf{Model 5}: For this model, the TotalSegmentator toolbox was used for reducing the number of unlabeled voxels by TotalSegmentator Pseudo Labeling.
            \item \textbf{Model 6}: To evaluate the effect of replacing the standard TCIA CT Lymph Node annotations with the according Bouget label refinements, a model was trained on this data combination. As a weakly supervised learning strategy, TotalSegmentator Pseudo Labeling was used. In the standard TCIA CT Lymph Node annotations, only enlarged LN components are contained, while the Bouget refinements include all sizes of LN components.
            \item \textbf{Model 7}: Finally, the NSCLC data were integrated into the training cohort. A model using TotalSegmentator Pseudo Labeling was trained on the data combination of the LNQ2023 training set, CT Lymph Nodes data with Bouget refinements, and the NSCLC datasets. The given tumor annotations of the NSCLC datasets were set to background class.
        \end{itemize}

        We evaluated the trained models on the 100 samples of the LNQ2023 test set and computed Dice score plus ASSD. The results of the experiments are shown in Table~\ref{tab:ablation_supervised_baseline_on_LNQ_testset}.

    \subsection{Performance Analysis regarding Lymph Node Shortest Diameter}
        \label{sec:shortest_diameter_analysis}

        To test whether a model solely trained on clinically enlarged pathological LN components was able to segment non-enlarged pathological LN components, we analyzed the predictions of \textbf{Model 5} and \textbf{Model 6} regarding their performance for different LN shortest diameters.

        To assess the performance, we iterated through all predicted LN components and computed the overlap of this predicted LN component on its ground truth annotation. The volume overlap was normalized by its volume, resulting in values between 0 and 1. This was then interpreted as a proxy of the model sensitivity. Thus, we were able to analyze the sensitivity over different LN sizes. We repeated the same assessment, iterated through all annotated ground truth LN components, and computed the overlap of this to the entire model prediction on the regarding volume, which was then a proxy for the precision.

\section{Results}
    The following presents an analysis of the preprocessing strategies, results of the ablation study, and an analysis of the segmentation performance on different LN component sizes.

    \subsection{Image and Annotation Preprocessing}

\begin{figure}
            \centering
     		\includegraphics[width=0.8\linewidth, interpolate]{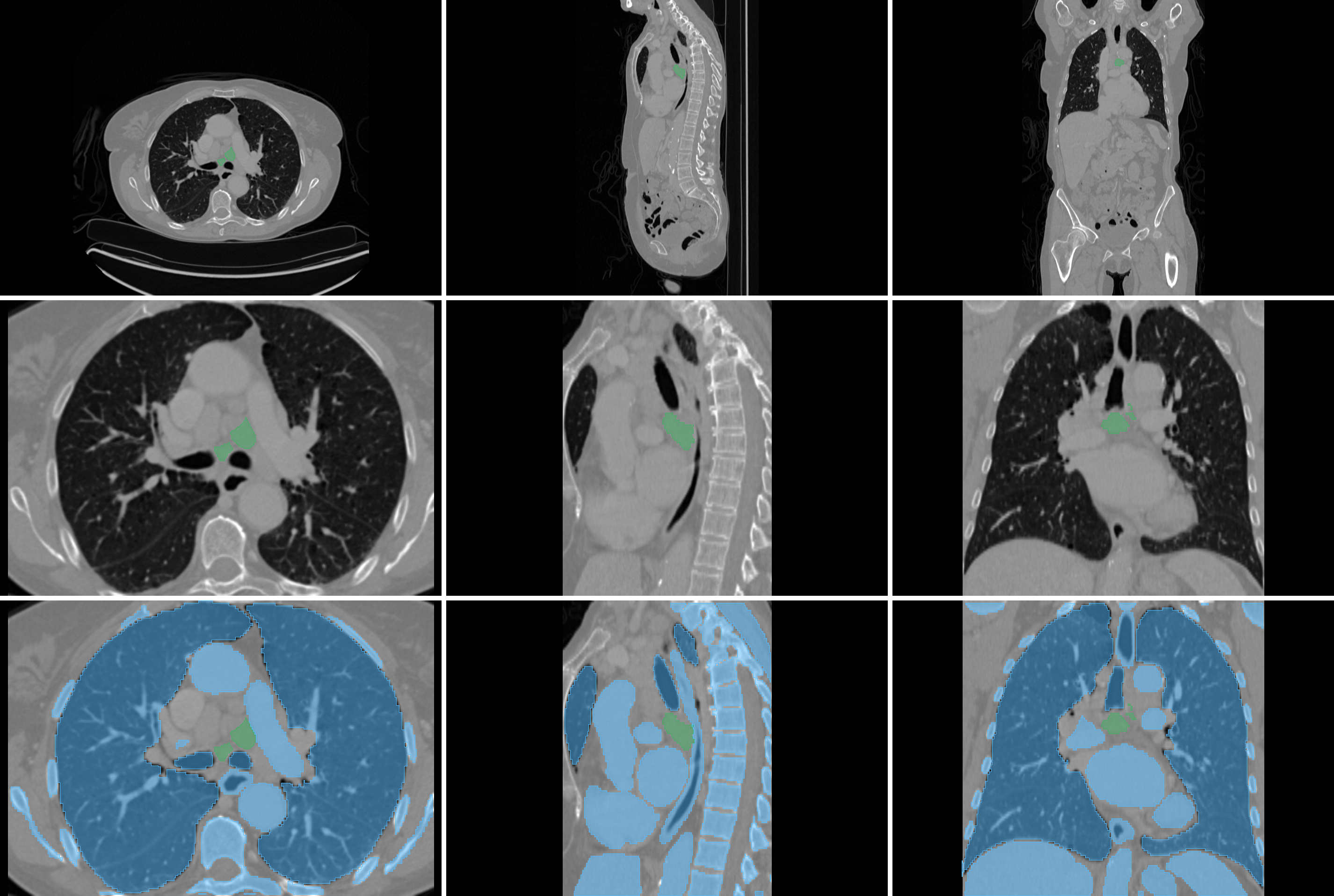}
          \caption{Preprocessing steps performed on LNQ2023 training set example, shown in three orthogonal views. \textbf{Top}: raw input volume with weak lymph node annotation (green), \textbf{Middle}: volume after lung bounding box cropping, \textbf{Bottom}: TotalSegmentator Pseudo Labeling setting known structures to the background (blue).}
          \label{Fig:totalseg_KD}
        \end{figure}

        First, the CT volumes were resampled to a common spacing of [3.0~mm, 0.93~mm, 0.93~mm]. The average number of voxels per CT volume and the ratio between labeled and unlabeled voxels are given in Table~\ref{tab:voxel_stats}. The number of voxels per CT volume was reduced from $38.2\cdot 10^{6}$ to $7.5\cdot 10^{6}$ with the ROI cropping, leading to an increase in the ratio of labeled to unlabeled voxels from $0.02\%$ to $0.10\%$. TotalSegmentator Pseudo Labeling further increased the number of labeled voxels, resulting in a ratio of $55.05\%$. Figure~\ref{Fig:totalseg_KD} shows the preprocessing steps of one LNQ2023 training set example.

        \begin{table}
            \centering
            \begin{tabular}{l | r | r  }
            Statistics Per CT Volume & Total Voxels [$\cdot 10^{6}$] & Labeled-Unlabeled Voxel Ratio [\%] \\
            \hline \hline 

            Raw Data & $38.2 \pm 24.5 $  & $0.02 \pm 0.03$ \\
            + ROI crop & $7.5 \pm 4.1 $ & $0.10 \pm 0.17$\\
            + TotalSegmentator PL  & $7.5 \pm 4.1 $ & $55.05 \pm 3.31 $ \\
            
           \end{tabular}
           \caption{Voxel statistics of the LNQ2023 training set for each preprocessing step.}
           \label{tab:voxel_stats}
        \end{table}

    \subsection{Ablation Study for Integration Strategy of Weakly Annotated Data}

                    \begin{figure}
            \centering
     		\includegraphics[width=0.8\linewidth]{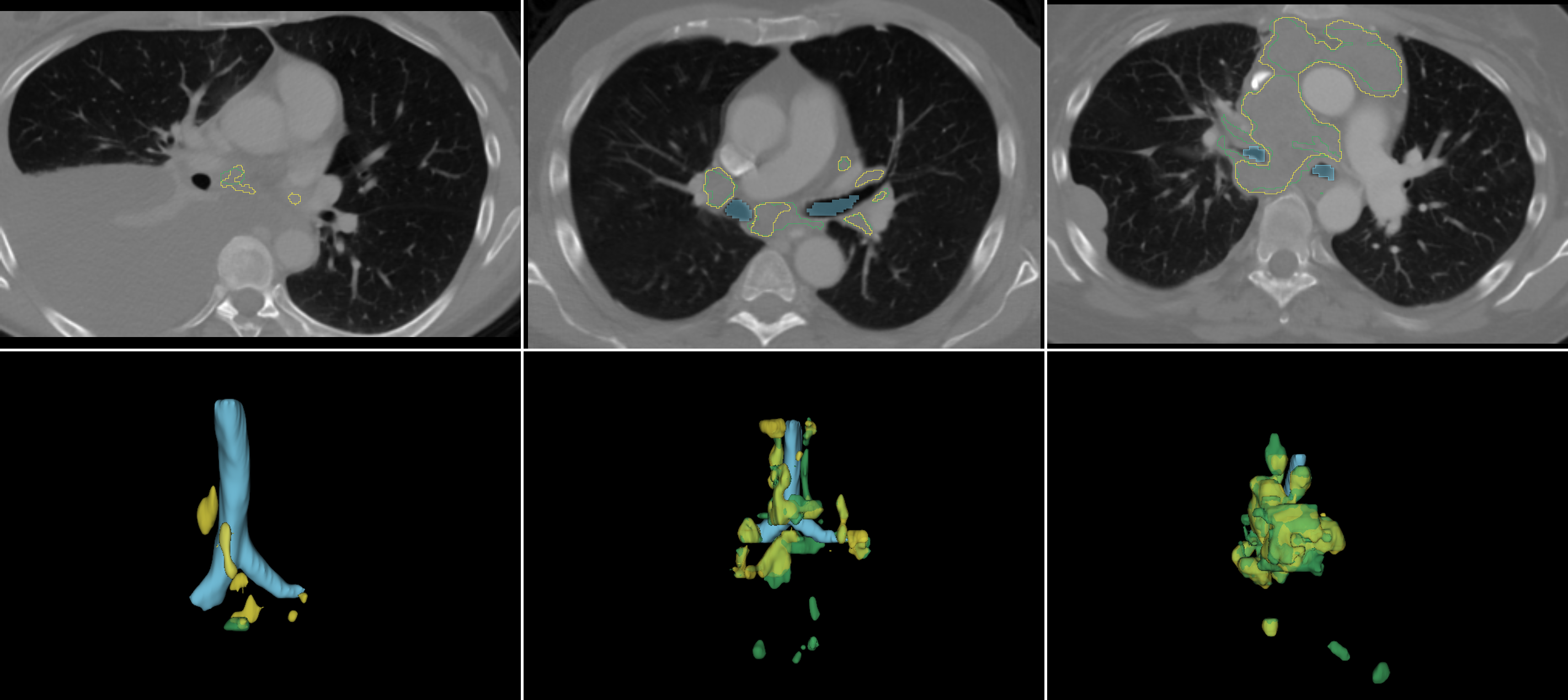}
          \caption{Different cases of LNQ2023 test set with ground truth annotation and model predictions. For intuitive visualization the trachea is shown in blue, model prediction is in yellow and ground truth in green. For inference \textbf{Model 7} was used. \textbf{Left}: worst case (Dice score 0.108, ASSD 19.2~mm), \textbf{Center}: average case (Dice score 0.626, ASSD 5.65~mm), \textbf{Right}: best case (Dice score 0.860, ASSD 2.38~mm)}
          \label{Fig:inference_examples}
        \end{figure}

        The results of the ablation study are presented in Table~\ref{tab:ablation_supervised_baseline_on_LNQ_testset}. \textbf{Model 1}, only trained on the fully annotated CT Lymph Nodes dataset containing only enlarged LN components, achieved a low Dice score of 0.172 and an ASSD of 48.95~mm on average. 

        Integration of the weakly annotated challenge training data improved the results overall for all applied weak annotation handling strategies. The worst performance was the noisy label training, with a Dice score of 0.343 and an ASSD of 9.37~mm. Loss masking and its variants, foreground instance coating, and TotalSegmentator Pseudo Labeling outperformed the noisy labels. Foreground instance coating did decrease the performance compared to raw loss masking in both metrics. The best weak annotation handling strategy was the TotalSegmentator Pseudo labeling with a Dice score of 0.601 and an ASSD of 7.08~mm.

                \begin{table}     
            \centering
            \begin{tabular}{l | r | r}
            Model & Dice Score $\uparrow$ & ASSD [mm] $\downarrow$ \\
            \hline \hline
            \textbf{Model 1}: CT Lymph Nodes dataset & $0.172 \pm 0.182$  & $48.95 \pm 64.49 $ \\
            \hline
            Integration of LNQ2023 training samples: & & \\
            \textbf{Model 2}: Noisy Label Strategy & $0.343 \pm 0.201 $ & $18.46 \pm 23.86$ \\
            \textbf{Model 3}: Loss Masking Strategy & $0.552 \pm 0.200 $ & $9.37 \pm 9.79$ \\
            \textbf{Model 4}: Foreground Instance Coating & $0.548 \pm 0.219 $ & $12.19 \pm 17.62$ \\
            \textbf{Model 5}: TotalSegmentator PL & $0.601 \pm 0.173$ & $7.08 \pm 7.54$ \\
            \hline
            Replace Annotations/Add data: & & \\
            \textbf{Model 6}: Bouget Label Refinements & $\textbf{0.665} \pm \textbf{0.143}$ & $4.47 \pm 4.67$ \\
            \textbf{Model 7}: Integration of NSCLC data & $0.663 \pm 0.136 $ & $\textbf{3.97}  \pm \textbf{2.83}$ \\
            
           \end{tabular}
           \caption{Performance of the lymph node segmentation model with different strategies to handle the weakly annotated data and additional training data. Models were tested on the 100 fully annotated samples of the LNQ2023 test set. Best achieved scores are highlighted in bold. ($\uparrow$: higher is better, $\downarrow$: lower is better) }
           \label{tab:ablation_supervised_baseline_on_LNQ_testset}
        \end{table}

        Changing the standard CT Lymph Nodes annotation to the Bouget refinements improved the performance. Important to note is that the Bouget refinements are annotations of all visible LNs. This gave an increase to a Dice score of 0.665 and an ASSD of 4.47~mm with the TotalSegmentator Pseudo Labeling. Furthermore, a model trained with the additional NSCLC datasets did have a similar Dice score and a slightly better ASSD. The significance of the segmentation performance is analysed with a non-parametric paired Wilcoxon signed-rank test in the Appendix~\ref{sec:appendix_significance}. Predictions of \textbf{Model 7} on three different test set cases of the LNQ2023 are shown in Figure~\ref{Fig:inference_examples}.

    \subsection{Performance Analysis regarding Lymph Node Shortest Diameter}

\begin{figure}
          \centering
            \includegraphics[]{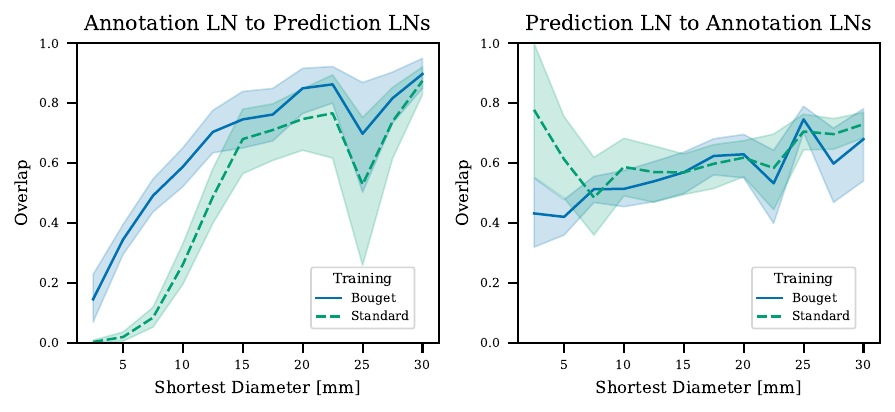}

          \caption{Overlap of each ground truth lymph node component with prediction and overlap of each predicted lymph node component with all ground truth lymph node components over the shortest diameter. Lymph node components were binned regarding shortest diameter in 2.5~mm steps. Model predictions were generated by \textbf{Model 5} (green) and \textbf{Model 6} (blue).}
          \label{Fig:overlap_between_pred_and_GT}
        \end{figure}

        In Figure~\ref{Fig:overlap_between_pred_and_GT}, the overlap between the ground truth LN component on prediction components and the overlap of predicted LN component on ground truth components is plotted. \textbf{Model 6}, which was trained on LN components of all sizes, achieved a better overlap between a ground truth LN component on a full model prediction on average for all different LN shortest diameters. \textbf{Model 5} did detect much fewer LN instances, which are shorter than 10~mm in diameter, than the \textbf{Model 6}.

        Overlap between predicted LN instances and complete ground truth annotations performs similarly, while for small LN instances, \textbf{Model 5} indicates a higher precision.

    \section{Challenge Submission}
    \label{sec:challenge_submission}
    
        Our submitted challenge model, trained via TotalSegmentator Pseudo Labeling on the combination of Bouget refined CT Lymph Node data, LNQ2023 training set, and the TCIA NSCLC datasets, ranked third place in the MICCAI LNQ2023 challenge. The model performed slightly worse on the test set than the developed models in the ablation study of Subsection~\ref{sec:ablation_studies_of_supervised_baseline}. We discuss reasons for this difference in Section~\ref{sec:discussion}. The nnUNet instance of the submission follows all the findings from the ablation study and refers to \textbf{Model 7} of the ablation study.

         \begin{table}[h]
            \centering
            \begin{tabular}{l | r | r}
            Authors (Team Name) & Dice Score $\uparrow$ & ASSD [mm] $\downarrow$ \\
            \hline \hline 
            \textbf{Rank 1}: Deissler et al. (Skeleton Suns) & $\textbf{0.674} \pm \textbf{0.165}$ & $4.5 \pm 4.7$ \\
            \textbf{Rank 2}: Zhang et al. (IMR) & $0.665 \pm 0.163$ & $5.4 \pm 4.3$ \\
            \textbf{Rank 3}: Fischer et al. (CompAI) & $0.628 \pm 0.193$ & $5.8 \pm 3.6$ \\
            \textbf{Rank 4}: Kondo et al. (HiLab) & $0.603 \pm 0.141$ & $8.2 \pm 11.8$ \\
            \textbf{Rank 5}: Engelson et al. (sofija\_engelson) & $0.569 \pm 0.185$ & $6.9 \pm 4.0$ \\
            \hline
            CompAI (ours) without Postprocessing & $0.660 \pm 0.150$ & $5.35 \pm 10.71$ \\
            \textbf{Model 7} & $0.663 \pm 0.136$ & $\textbf{3.97} \pm \textbf{2.83}$ \\
            
           \end{tabular}
           \caption{Performance of the top five performing team models on the 100 samples from LNQ2023 test set. Additionally, the performance of our submitted model without the removal of small LN components is given, as well as the \textbf{Model 7} from the ablation study. Best achieved scores are highlighted in bold.}
           \label{tab:score_on_test_set}
        \end{table}

        For the challenge submission, the default nnUNet normalization scheme was replaced by intensity clipping to $[-150,350]$ inspired by the work of \cite{bouget2019semantic} and intensity standardization. Furthermore, the challenge submission model was trained with a learning rate of $1e-3$ instead of the default $1e-2$ in order to stabilize the training process. Thus, we increased the number of epochs from default 1000 epochs to 2000 epochs. To train one nnUNet instance the full available data were used, also integrating the missing 20 samples of CT Lymph Node for original validation.

        The prediction was postprocessed by a connected component analysis for the challenge submission. Predicted LN components with a short diameter smaller than 9.5~mm were removed from the prediction so that only LN components that are considered enlarged are kept. The ranking of the challenge participating teams is given in Table~\ref{tab:score_on_test_set}. Our submission was ranked third place in the Lymph Node Quantification Challenge 2023 with a Dice score of 0.628 and an ASSD of 5.8~mm. 

        \textbf{Model 7} of our ablation study would have achieved third place in the Dice score while having the lowest standard deviation. Furthermore, it would have scored the best ASSD among all challenge submission models.

    \section{Discussion}
    \label{sec:discussion}
    For the challenge, only semantic segmentation maps have been provided, while the CT Lymph Nodes dataset comes with instance segmentation annotations. Therefore, original annotations of the CT Lymph Nodes were interpreted as semantic segmentation maps, and LN components were created by a connected component analysis. Thus, there is a difference in the reported LN instances in the work of \cite{roth2015tcia} compared to ours.

    In Table~\ref{tab:LNs_Stats_Per_Dataset}, the number of LN components is provided for each dataset. For the CT Lymph Nodes dataset, which should only contain enlarged LNs, also a subset of non-enlarged LNs are included. An explanation for this trend is that clinicians follow the RECIST guidelines and thus only consider axial slice directions. For the LNQ2023 training set, the same trend holds but is less prominent. This might originate from the weak annotating done by the physicians, probably resulting in a bias toward the larger pathological LNs per image volume. 

    Another related surprising finding was the difference in enlarged LN instances in CT Lymph Nodes and the Bouget refinements. There might be a bias originating from the intention of annotating all visible LNs in the refinements, while the clinicians searched only for enlarged LNs via the RECIST criteria in the standard dataset. Another factor is that each LN of a neighboring LN cluster might be considered as healthy, while annotating all of them and processing the cluster with a connected component analysis can result in an enlarged LN component.

    The TCIA CT Lymph Nodes dataset was essential for the development of the proposed method, as it provided fully annotated training cases. A possible solution to solve the task with only the weakly annotated challenge data is to follow the work of \cite{nguyen2020learning}, which refers to foreground instance coating. By using TotalSegmentator Pseudo Labeling the performance in this scenario would improve. Another approach would be noisy label training. We did not perform any experiments on the challenge data alone.

    In our scenario, the noisy label training improved the performance compared to only using CT Lymph Nodes. The integration of the LNQ2023 training set outweighed the effect of false negatives generated from unlabeled LN instances.
    
    We hypothesize the reason for the failure of the foreground coating is the difficulty of labeling LNs in a binary manner, as LN are known to be confluent and miss sharp intensity drops at their boundary. Thus, the instance coating might generate a lot of false negatives. Increasing the margin of the coating might reduce this issue and lead to performance gains. Furthermore, the presence of bulky lymph nodes in which only one LN instance was annotated will result in false negative annotations.

    TotalSegmentator Pseudo Labeling offers limited generalization to other tasks, as it is only applicable to CT modality and differs in coverage for all body regions. Furthermore, there is also a small overlap between LN voxels and TotalSegmentator structures, leading to false negatives. There might be better structure subsets of the TotalSegmentator for the LN segmentation task.

    The segmentation models benefit from integrating the Bouget refinements, also reported in the work of \cite{bouget2021mediastinal}. We support their hypothesis that the inclusion of all visible LNs is an efficient form of data augmentation, integrating possible LN locations. Another aspect is the incorporation of non-enlarged LNs into the training, which is essential as models trained only on enlarged LNs are not generalizing well on small LNs. 

    We hypothesize that the better performance of ASSD and more stable Dice scores, when including the NSCLC data, originates from the higher number of training samples and the learning of lung cancer anomalies as background. 

    Extending the method with a semi-supervised learning method, in the form of Totalsegmentator Pseudo Labeling, was shown to improve the model. Additionally, pseudo labeling of the remaining unlabeled voxels as described in the work of \cite{huang2022revisiting} might further increase the performance.

    Until the final challenge submission, the goal of the challenge, that all pathological lymph nodes should be segmented, was ambiguous to us. Our intention was to segment only clinically enlarged lymph nodes. Thus, we introduced the postprocessing of filtering the segmentation regarding LN enlargement. This postprocessing led to a lower performance. Nevertheless, we were still able to achieve third place in the final challenge ranking.
    
    In this work, different strategies to handle weak annotations were proposed that significantly improved the performance on the task of mediastinal pathological lymph node segmentation. The usage of the TotalSegmentator was highly beneficial in our case as ROI cropping but also as a network providing informative pseudo labels. Different datasets were integrated into the training and were fundamental for the submission model. One important finding is that the integration of non-pathological lymph nodes also aided our task of pathological lymph node segmentation.

%%%%%%%%%%%%%%%%%%%%%%%%%%%%%%%%%%%%%%%%%%%%%%%%%%%%%%%%%%%%%%%%%%%%%%%
% Mandatory Sections. Please complete, especially for final publication
%%%%%%%%%%%%%%%%%%%%%%%%%%%%%%%%%%%%%%%%%%%%%%%%%%%%%%%%%%%%%%%%%%%%%%%

% Acknowledgements.
% Please include any funding, intellectual contributions not included in the authorship, and any other acknowledgements.
\acks{Stefan Fischer has received funding by the Deutsche Forschungsgemeinschaft (DFG, German Research Foundation) – 515279324  / SPP 2177. Johannes Kiechle was supported by the DAAD programme Konrad Zuse School of Excellence in reliable Artificial Intelligence (relAI), sponsored by the Federal Ministry of Education and Research.}

% Ethical Standards.
% Please edit with the appropriate ethics considerations for your work. Include any pertinent IRB information, etc.
%
% Please note that the submission requirements included:
% The work presented must follow appropriate ethical standards in conducting research and writing the manuscript, following all applicable laws and regulations regarding treatment of animals or human subjects.
\ethics{The work follows appropriate ethical standards in conducting research and writing the manuscript, following all applicable laws and regulations regarding treatment of animals or human subjects.}

% Conflict of Interest
% Declaration of possible conflicts of interest: Authors must disclose any financial, organisational, commercial or personal conflicts of interest that might bias their work.
% If no conflicts, please say "We declare we don't have conflicts of interest."
\coi{We declare we do not have conflicts of interest.}

% Data availability
\data{All experiments and models are performed and trained on publicly available data. The LNQ2023 challenge data will be published by the challenge hosts in the future as part of TCIA. Bouget refinements are accessible via the Google-Drive links via \\ \href{https://github.com/dbouget/ct_mediastinal_structures_segmentation}{https://github.com/dbouget/ct\_mediastinal\_structures\_segmentation}.
}

\bibliography{sample}

% Manual newpage inserted to improve layout of sample file - not
% needed in general before appenDices.
% \newpage

% Appendix is optional
\clearpage
\appendix

\begin{landscape}
\section{Significance Testing on Ablation Study}
\label{sec:appendix_significance}
	For the ablation study of Subsection~\ref{sec:ablation_studies_of_supervised_baseline} p-values are computed via the paired sample non-parametric Wilcoxon signed-rank test. The results are given in Table~\ref{tab:pvalue_dsc} and Table~\ref{tab:pvalue_assd}.
\vspace{\fill}

          \begin{table}[h]
          \label{tab:pvalues}
          \footnotesize
 
\centering
            \begin{tabular}{l | l | l | l | l | l | l | l | l}
            & \textbf{Model 1} & \textbf{Model 2} & \textbf{Model 3} & \textbf{Model 4} & \textbf{Model 5} & \textbf{Model 6} & \textbf{Model 7} & \textbf{Dice Score} \\
            \hline \hline 
            \textbf{Model 1} & - & $1.5e^{-10}$ & $3.6e^{-17}$ & $5.0e^{-17}$ & $5.9e^{-18}$ & $5.9e^{-18}$ & $4.0e^{-18}$ & $0.172 \pm 0.182$ \\
            \textbf{Model 2} & $1.5e^{-10}$ & - & $3.2e^{-13}$ & $1.4e^{-13}$ & $5.0e^{-17}$ & $2.0e^{-17}$ & $2.4e^{-17}$ & $0.343 \pm 0.201 $ \\
            \textbf{Model 3} & $3.6e^{-17}$ & $3.2e^{-13}$ & - & $0.92$ & $5.8e^{-4}$  & $4.3e^{-12}$ & $1.0e^{-10}$ & $0.552 \pm 0.200 $ \\
            \textbf{Model 4} & $5.0e^{-17}$ & $1.4e^{-13}$ & $0.92$ & - & $1.1e^{-5}$ & $3.2e^{-11}$ & $1.2e^{-10}$  & $0.548 \pm 0.219 $ \\
            \textbf{Model 5} & $5.9e^{-18}$ & $5.0e^{-17}$ & $5.8e^{-4}$ & $1.1e^{-5}$ & - & $1.1e^{-10}$ & $2.2e^{-11}$ & $0.601 \pm 0.173$ \\
            \textbf{Model 6} & $5.9e^{-18}$ & $2.0e^{-17}$ & $4.3e^{-12}$ & $3.2e^{-11}$ & $1.1e^{-10}$ & - & $2.5e^{-1}$ & $0.665 \pm 0.143$ \\
            \textbf{Model 7} & $4.0e^{-18}$ & $2.4e^{-17}$ & $1.0e^{-10}$ &  $1.2e^{-10}$ & $2.2e^{-11}$ & $2.5e^{-1}$ & -  & $0.663 \pm 0.136 $ \\
            \textbf{Dice Score} & $0.172 \pm 0.182$ & $0.343 \pm 0.201 $ & $0.552 \pm 0.200 $ & $0.548 \pm 0.219 $ & $60.1 \pm 0.173$ & $0.665 \pm 0.143$ & $0.663 \pm 0.136 $ & - \\
            
           \end{tabular}
           \caption{P-Value of non-parametric paired Wilcoxon signed-rank test of null hypothesis for Dice score.}
           \label{tab:pvalue_dsc}
        \end{table}

                \begin{table}[h]
 
            \centering
            \begin{tabular}{l | l | l | l | l | l | l | l | l}
            & \textbf{Model 1} & \textbf{Model 2} & \textbf{Model 3} & \textbf{Model 4} & \textbf{Model 5} & \textbf{Model 6} & \textbf{Model 7} & \textbf{ASSD [mm]} \\
            \hline \hline 
            \textbf{Model 1} & - & $3.9e^{-8}$ & $2.9e^{-16}$ & $4.2e^{-15}$ & $4.0e^{-18}$ & $3.9e^{-18}$ & $3.9e^{-18}$ & $48.95 \pm 64.49 $ \\
            \textbf{Model 2} & $3.9e^{-8}$ & - & $7.7e^{-10}$ & $6.6e^{-10}$ & $6.7e^{-17}$ & $1.6e^{-17}$ & $7.3e^{-18}$ & $18.46 \pm 23.86$ \\
            \textbf{Model 3} & $2.9e^{-16}$ & $7.7e^{-10}$ & - & $2.0e^{-3}$ & $7.3e^{-4}$ & $5.7e^{-13}$ & $1.3e^{-14}$ & $9.37 \pm 9.79$ \\
            \textbf{Model 4} & $4.2e^{-15}$ & $6.6e^{-10}$ & $2.0e^{-3}$ & - & $3.2e^{-7}$ & $2.6e^{-14}$ & $1.5e^{-15}$ & $12.19 \pm 17.62$\\
            \textbf{Model 5} & $4.0e^{-18}$ & $6.7e^{-17}$ & $7.3e^{-5}$ & $3.2e^{-7}$ & - & $7.4e^{-11}$ & $1.6e^{-13}$   & $7.08 \pm 7.54$ \\
            \textbf{Model 6} & $3.9e^{-18}$ & $1.6e^{-17}$ & $5.7e^{-13}$ & $2.6e^{-14}$ & $7.4e^{-11}$ & - & $7.8e^{-1}$  & $4.47 \pm 4.67$\\
            \textbf{Model 7} & $3.9e^{-18}$ & $7.3e^{-18}$ & $1.3e^{-14}$ &  $1.5e^{-15}$ & $1.6e^{-13}$ & $7.8e^{-1}$ & -  & $3.97  \pm 2.83$ \\
            \textbf{ASSD [mm]} & $48.95 \pm 64.49 $ & $18.46 \pm 23.86$ & $9.37 \pm 9.79$ & $12.19 \pm 17.62$ & $7.08 \pm 7.54$ & $4.47 \pm 4.67$ & $3.97  \pm 2.83$ & - \\
            
           \end{tabular}
           \caption{P-Value of non-parametric paired Wilcoxon signed-rank test of null hypothesis for ASSD.}
           \label{tab:pvalue_assd}
        \end{table}
        \end{landscape}

\end{document}